\documentclass[journal]{IEEEtran}

\usepackage{times}
\usepackage{graphicx}
\usepackage{algorithm}

\usepackage{algpseudocode}
\usepackage{amsmath}

\usepackage[numbers,sort&compress]{natbib} 
\usepackage{multirow}

\usepackage{epstopdf}

\usepackage{amssymb}

\usepackage{xcolor}

\usepackage{url}

\usepackage{subfigure}

\usepackage{bm}

\newcommand{\rev}[1]{{\color{black}#1}}

\usepackage{enumerate}

\ifCLASSINFOpdf
\else
\fi

\begin{document}
%
\title{Learning View-Specific Deep Networks for Person Re-Identification}

\author{\IEEEauthorblockN{Zhanxiang Feng, Jianhuang Lai, and Xiaohua Xie}
\thanks{Corresponding author: Jianhuang Lai (e-mail: stsljh@mail.sysu.edu.cn).

Zhanxiang Feng is with the School of Electronics and Information Technology, Sun Yat-Sen University, Guangzhou 510006, China, and with the Xin Hua College of Sun Yat-Sen University, Guangzhou 510006, China (e-mail: fengzhx@mail2.sysu.edu.cn)

Jianhuang Lai and Xiaohua Xie are with the School of Data and Computer Science, Sun Yat-Sen University, Guangzhou 510006, China, and with the Guangdong Key Laboratory of  Machine  Intelligence and Advanced Computing, Ministry of Education, Guangzhou 510006, China (e-mail: stsljh@mail.sysu.edu.cn and xiexiaoh6@mail.sysu.edu.cn).
}}

\markboth{IEEE TRANSACTIONS ON IMAGE PROCESSING}%
{Shell \MakeLowercase{\textit{et al.}}: Bare Demo of IEEEtran.cls for IEEE Transactions on Magnetics Journals}

\maketitle

\begin{abstract}
In recent years, a growing body of research has focused on the problem of person re-identification (re-id). The re-id techniques attempt to match the images of pedestrians from disjoint non-overlapping camera views. A major challenge of re-id is the serious intra-class variations caused by changing viewpoints. To overcome this challenge, we propose a deep neural network-based framework which utilizes the view information in the feature extraction stage. The proposed framework learns a view-specific network for each camera view with a cross-view Euclidean constraint (CV-EC) and a cross-view center loss (CV-CL). We utilize CV-EC to decrease the margin of the features between diverse views and extend the center loss metric to a view-specific version to better adapt the re-id problem. Moreover, we propose an iterative algorithm to optimize the parameters of the view-specific networks from coarse to fine. The experiments demonstrate that our approach significantly improves the performance of the existing deep networks and outperforms the state-of-the-art methods on the VIPeR, CUHK01, CUHK03, SYSU-mReId, and Market-1501 benchmarks.
\end{abstract}


\IEEEpeerreviewmaketitle

\section{Introduction}



\IEEEPARstart{W}{ith} the increasing ubiquity of closed-circuit television cameras, the problem of person re-identification (re-id) has attracted considerable research attention. The purpose of re-id techniques is to match the images of pedestrians from disjoint camera views. The application of re-id techniques in intelligent video surveillance systems will be beneficial to enhance security in public areas. Conducting re-id over disjoint non-overlapping camera views is challenging because of the dramatic visual changes caused by variations in illumination, image quality, and especially viewpoint.

\begin{figure}[t]
  \centering
  \includegraphics[width=0.48\textwidth]{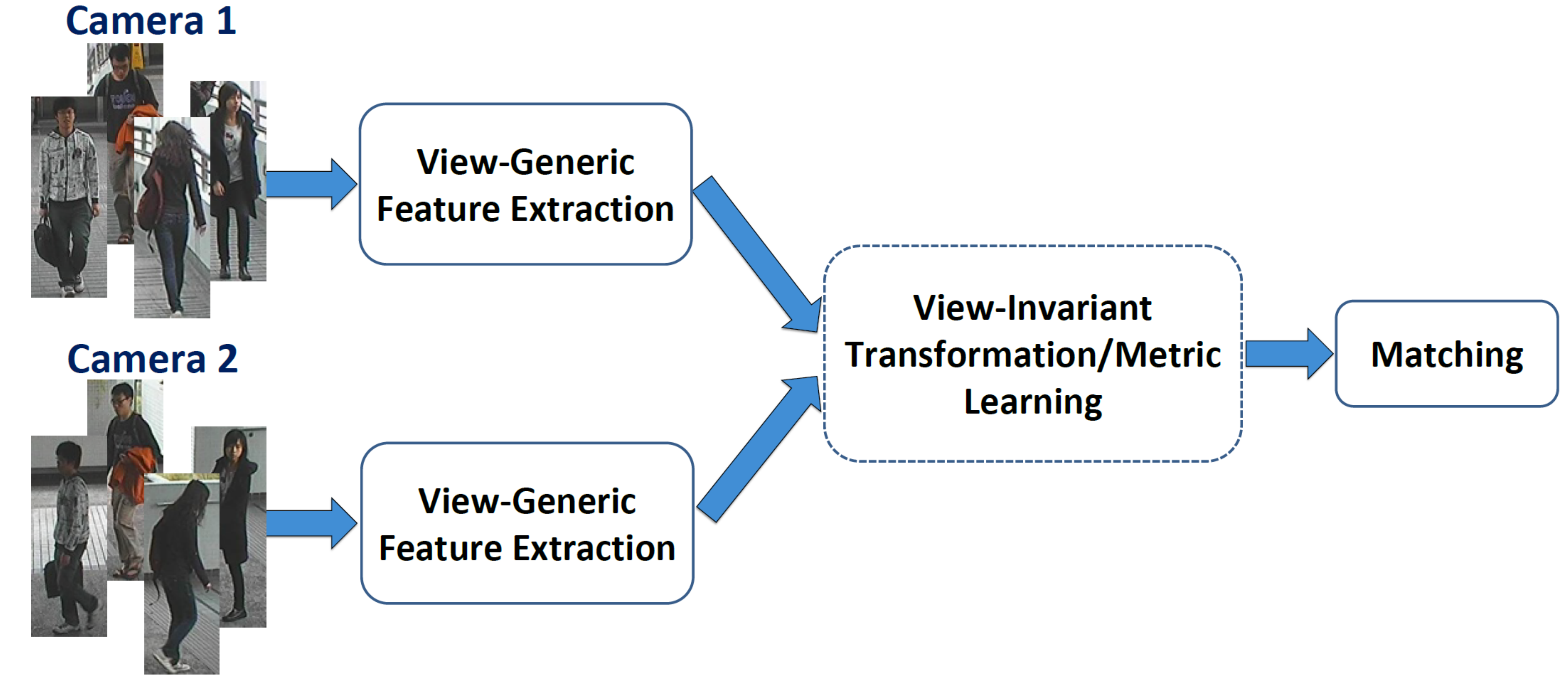}
  \caption{Traditional re-id framework for addressing the cross-view problem. Traditional re-id framework conducts the same implementation to extract view-generic features for different camera views. The feature extraction process is usually followed by a view-invariant discriminative transformation or metric, which is shared across the changing viewpoints in most cases. The dashed window indicates that the step can be skipped.}\label{TIP:Traditional_framework}
\end{figure}

In the existing literature, most re-id methods consist of a feature extraction process and a view-invariant discriminative transformation or metric \cite{koestinger2012large,chen2016similarity,zheng2013reidentification,pedagadi2013local,xiong2014person,liao2015person,paisitkriangkrai2015learning}. Figure \ref{TIP:Traditional_framework} displays a traditional re-id framework for addressing the cross-view problem. Conventional re-id approaches first extract view-generic features for different views. Then, a view-invariant model is learned to narrow the gaps between the images of the intra-class pedestrians and increase the distances between the inter-class samples. Three issues prevent traditional methods from solving the re-id problem. First, traditional features characterize pedestrians without any view-specific knowledge. View-generic features are inadequate for solving a re-id problem when dramatic changes in appearance occur between disjoint camera views. Second, feature extraction and view-invariant model learning are independent. During training, the view-invariant model propagates no information back to the view-generic features. View information, which is commonly exploited by discriminative model learning algorithms, is rarely utilized to improve the feature extraction process. Learning a model with excellent generalizability simply by metric learning is difficult because the view-generic features of the same person vary across different views. Third, traditional methods learn a view-invariant model shared by all views. Utilizing a shared model for all views ignores the discrepancies among different views. Compared with the shared model, view-specific models can cover stronger view-related information. With such information, view-specific models can achieve better performance than view-generic ones. However, the amount of computational resources grows dramatically with the increasing number of camera views. Currently, few studies have focused on the learning of view-specific models.

Although view information may be useful for improving the performance of the existing methods, learning view-specific features or transformations for re-id is still under study. The current literature focuses on learning view-invariant but view-generic discriminative information. The existing approaches mainly integrate view information into metric learning methods by projecting the view-generic features onto a view-invariant common space \cite{jia2016cross,li2013locally,wang2016zero,mignon2012pcca}. To further exploit view-related information, some recent works \cite{he2016cross,chen2017person,chen2016asymmetric,li2015cross,wang2014cross} attempt to learn a view-specific dictionary or transformation for each view and project the view-generic features onto the corresponding view-specific spaces. Although view-specific models can minimize the discrepancies among changing viewpoints, the process of view-generic feature extraction ignores the view information hidden in the low-level features. The using of view-specific features may improve the performance of the existing frameworks. Nevertheless, few studies have explored the topic of view-specific feature extraction for re-id issue.

Data-driven features are more effective than empirically designed features in exploiting view-specific information. Deep learning is the most effective technique to learn discriminative features from the training data. Deep networks have been successfully applied in various computer vision tasks, including image classification \cite{krizhevsky2012imagenet,he2016deep}, face recognition \cite{hu2014discriminative,sun2014deep}, and action recognition \cite{wang2015action,kong2017deeply}. In the last few years, deep learning-based approaches have been proved to be effective for re-id \cite{yi2014deep,li2014deepreid,ahmed2015improved,ding2015deep,varior2016siamese,varior2016gated,chen2016deep,su2016deep,cheng2016person,liu2017end,wang2014learning,ustinova2016learning}.
Deep models may exploit the view information and extract discriminative view-related features by using a back propagation algorithm. However, the existing approaches share the parameters of deep networks across all views, particularly for the layers that extract low-level features. Therefore, view-specific information may be ignored during feature extraction stage.

According to the above discussion, we propose to extract view-specific features by using deep networks. To the best of our knowledge, our approach is the first to extract view-specific features by using deep networks for re-id. To narrow the gaps between features from changing viewpoints, we integrate a cross-view Euclidean constraint (CV-EC) into the proposed framework. The goal of CV-EC is to decrease the distances between the deep features of the same person from disjoint camera views. Together with CV-EC, we integrate another loss called the cross-view center loss (CV-CL) to improve the discriminative ability of the view-specific deep networks. Center loss has been proved to be effective for face recognition \cite{wen2016discriminative} by narrowing the margin between the samples and their corresponding class centers. In this paper, we extend the center loss to a view-specific version that better fits the re-id problem. Moreover, we propose an iterative optimization algorithm (ICV-ECCL) to learn CV-EC and CV-CL alternatively and optimize the parameters of the view-specific networks from coarse to fine. Finally, we extend CV-EC and CV-CL to a multi-view version to deal with the application using more than two cameras.

The study makes the following contributions.
\begin{itemize}
\item We propose a novel framework to learn view-specific networks for feature extraction in re-id and demonstrate that the utilization of view-specific information is crucial for extracting robust re-id representations.
\item We propose the CV-EC and CV-CL constraints to integrate view information into view-specific deep networks and minimize the distances between the features across varying views. The proposed framework is adaptive to any baseline network and benchmark.
\item We propose an iterative algorithm (ICV-ECCL) to optimize the parameters of view-specific networks from coarse to fine.
\item We evaluate the proposed framework by extensive comparisons between ICV-ECCL and a variety of methods on the VIPeR, CUHK01, CUHK03, SYSU-mREID, and Market-1501 benchmarks. The experimental results validate that the proposed framework significantly improves the performance of the existing deep networks and outperforms the state-of-the-art methods.
\end{itemize}

The rest of this paper is organized as follows: In Section \uppercase\expandafter{\romannumeral2}, we review the related works. We then introduce the overall framework and the optimization algorithm in Section \uppercase\expandafter{\romannumeral3}. Section \uppercase\expandafter{\romannumeral4} presents the experimental results of the comparisons and the self-evaluation. Section \uppercase\expandafter{\romannumeral5} concludes the paper.

\section{Related Work}

\subsection{View-Generic Features for Re-Id}
View-generic features are widely used for re-id. The existing works extract view-generic features to describe the visual characteristics of the objective pedestrians captured in disjoint cameras. The color \cite{zhao2013unsupervised,shen2015person,bazzani2010multiple,cheng2011custom,yang2014salient,matsukawa2016hierarchical}, shape \cite{belongie2002shape,wang2007shape,weinberger2009distance}, texture \cite{farenzena2010person,liu2014semi,shi2015transferring}, and spatio-temporal features \cite{gheissari2006person,bedagkar2012part} are the most commonly used characteristics to describe the target pedestrian regions. Descriptors such as LBP \cite{liu2014semi}, HOG \cite{schwartz2009learning,wang2007shape}, Gabor \cite{li2013locally,ma2012bicov}, ELF \cite{gray2008viewpoint}, and SIFT \cite{zhao2013unsupervised,park2006vise} have been successfully applied to match persons from disjoint views. Researchers have also extracted global features \cite{shen2015person,cheng2011custom}, local features \cite{prosser2010person,ma2013domain}, and patch-based features \cite{bazzani2013symmetry,dikmen2011pedestrian} to integrate structural constraints into view-generic features. Moreover, some studies combine multiple features to form better descriptors \cite{kawai2012person,shi2015person}. In addition to hand-crafted features, some unsupervised learning-based view-generic representations have been used for re-id, including sparse coding \cite{mignon2012pcca,he2016cross}, bag-of-words \cite{van2009learning,matsukawa2016hierarchical}, and fisher vector \cite{ma2012local}. Moreover, saliency \cite{yang2014salient,zhao2013unsupervised} and pedestrian attributes \cite{layne2012person,liu2012attribute,su2015multi,layne2012towards} have been used to match people across different cameras.
\subsection{View-Invariant Transformations and Metrics for Re-Id}
The learning of view-invariant transformations and metrics for re-id is of great interest to researchers. View-invariant transformation and metric learning algorithms can exploit label information and narrow the gaps between different views. In the existing literature, supervised learning-based approaches demonstrate superior performance to that of unsupervised learning-based approaches. KISSME \cite{koestinger2012large}, LMNN \cite{weinberger2009distance}, SCSP \cite{chen2016similarity}, RankSVM \cite{prosser2010person}, PCCA \cite{mignon2012pcca}, LFDA \cite{pedagadi2013local}, and ITML \cite{davis2007information} have been proposed to learn discriminative view-invariant representations.

In addition to label information, view information has been used to learn robust re-id representations. Traditionally, researchers project the features from both camera views onto a common space by a shared view-invariant model \cite{jia2016cross,li2013locally,wang2016zero,mignon2012pcca}. In recent years, some works have attempted to learn view-specific transformations or models \cite{he2016cross,chen2016asymmetric,li2015cross,chen2017person,wang2014cross}. Further, view-specific models have been proved to be superior to the shared models regarding the generalizability across the appearance variations caused by viewpoint changes. Chen et al. \cite{chen2017person} proposed a view-specific re-id framework by the feature augmentation of different views. The framework executed a view-specific learning algorithm to measure the camera correlations and transform the features from disjoint views to a new adaptive space by adaptive augmentation. Chen et al. \cite{chen2016asymmetric} also introduced an asymmetric distance model for cross-view feature mapping to extract discriminative features against the changing viewpoints. The asymmetric distance model learns camera-specific projections to transform low-level features from both views to a common space. Li et al. \cite{li2015cross} proposed a cross-view projective dictionary learning-based method, which learned a specific dictionary for each view, to cope with the re-id task.
\subsection{Deep Learning Models for Re-Id}
Deep learning is playing an increasingly significant role for the re-id task.
The existing deep network-based methods focus on designing Siamese networks that combine the input pairs of pedestrians from multiple camera views \cite{yi2014deep,li2014deepreid,ahmed2015improved,ding2015deep,varior2016siamese,varior2016gated}. For example, Yi et al. \cite{yi2014deep} and Li et al. \cite{li2014deepreid} used a Siamese neural network to optimize the parameters of deep networks with pairs of inputs. Ahmed et al. \cite{ahmed2015improved} further extended the Siamese neural network by embedding a layer to compute the cross-input neighborhood differences. Varior et al. \cite{varior2016siamese} jointly combined a Siamese neural network with LSTM to recurrently extract spatial-structured features. Varior et al. also proposed a Gated S-CNN \cite{varior2016gated} to match pedestrians from disjoint views horizontally.

Another common strategy for learning view-invariant deep features is to learn a discriminative cross-view metric through the ensemble layers \cite{chen2016deep,su2016deep,cheng2016person,liu2017end,wang2014learning}. Chen et al. \cite{chen2016deep} proposed a deep ranking model to improve the ranking of the correct match of probe images with a learning-to-rank algorithm. Su et al. \cite{su2016deep} proposed a semi-supervised deep attribute learning-based method for re-id. They first trained a deep network with labeled attributes in an extra dataset and then fine-tuned the network with the re-id dataset by using the attribute triplet loss. Liu et al. \cite{liu2017end} proposed a novel soft attention-based model called the end-to-end comparable attention network (CAN). The CAN model stimulates human perception in judging whether the parts are from the same person or not.
\subsection{Conclusion on Current Approaches}
View information needs to be further exploited by the existing re-id approaches. On one hand, the existing traditional re-id frameworks utilize view information simply by using view-invariant transformations or metrics. The main drawback of the view-specific model-based methods lies in the process of feature extraction. For most cases, the existing methods conduct the same implementation to extract view-generic features from different views, which weakens the strength of the view-related information during feature extraction and deteriorates the performance of the overall framework. To overcome the drawback of view-specific models and improve their generalizability, we need to extract view-specific discriminative features.

On the other hand, few works pay attention to learn view-specific deep features for re-id in an end-to-end manner. Most deep learning-based methods share the parameters of a single network for disjoint views. Such approaches learn the view-generic information while neglecting the view-specific information. The learning of an individual deep network for each camera view can cover more view-specific information and improve the generalizability towards viewpoint changes. On the basis of the above observations, we focus on training view-specific deep networks to extract view-related features, which is crucial for improving the generalizability of the re-id models.
\begin{figure*}[thb]
  \centering
  \includegraphics[width=0.96\textwidth]{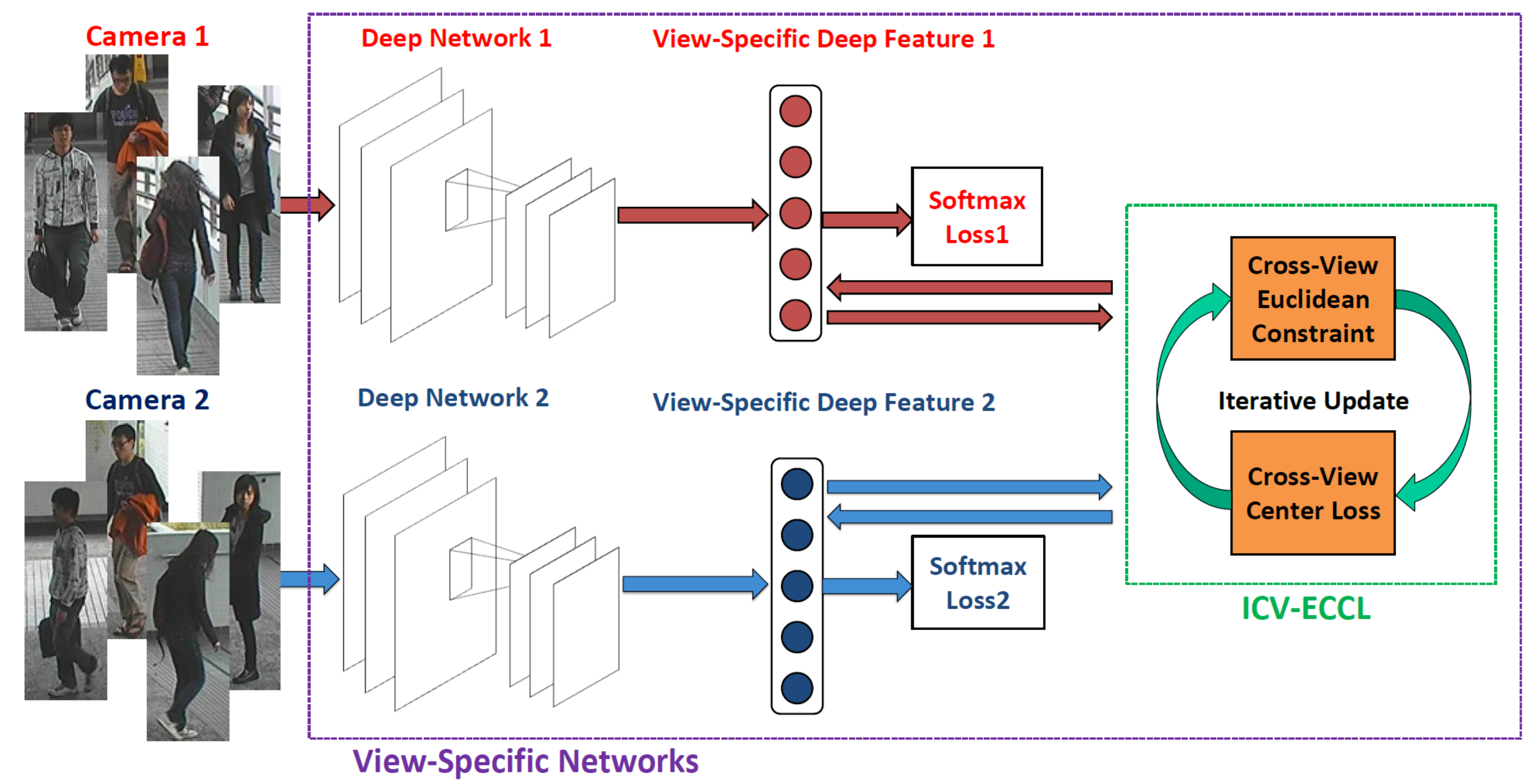}
  \caption{Proposed re-id framework. View-specific deep networks are trained to extract view-related features. The cross-view Euclidean constraint and the cross-view center loss are integrated into the view-specific networks to narrow the gaps between the features from changing viewpoints, and they are employed in an iterative manner.}\label{2}
\end{figure*}
\section{Proposed Framework}
\subsection{Overview}
By sharing the parameters of the lower layers, the existing works extract the view-generic low-level features for changing viewpoints. View information is thus ignored during low-level feature extraction. To solve this problem, we propose to learn view-specific deep networks to extract discriminative view-related features for re-id. Figure 2 illustrates the proposed framework. Our framework is different from conventional deep models in that it learns different deep networks for different views. Further, we integrate CV-EC into the framework to align the deep features from disjoint views. We also introduce CV-CL to narrow the gaps between the features across different views. Furthermore, CV-EC and CV-CL are utilized iteratively (ICV-ECCL) to update the parameters of view-specific networks. Finally, we extend CV-EC and CV-CL to a multi-view version.
\subsection{Cross-View Euclidean Constraint}
As shown in Figure 2, a view-specific network is learned for each view in the proposed framework. We seek to learn the view-related features during training. For view-specific deep networks, the cross-view intra-class distance may be very large without the consideration of any cross-view constraint. We need to minimize the cross-view intra-class distances between pairs of view-specific features. To address this problem, we integrate a constraint into the view-specific networks to guarantee that the features of the same people under disparate views are as close as possible.

In this section, we introduce the main principle of CV-EC. To simplify the discussion, we only consider the situation of two cameras. In fact, the proposed method can be extended to multiple cameras. Given the deep features from disjoint views as $\{\bm{x}^1_{ip},\bm{x}^2_{iq}\},1\le i\le M, 1\le p\le K^1_i, 1\le q\le K^2_i$, where $i$ denotes the identity of the pedestrians, $M$ represents the number of identities, $(ip,iq)$ refer to the $p$th and $q$th feature of the $i$th identity from view 1 and view 2, respectively, and $K^1_i$ and $K^2_i$ denote the numbers of the $i$th identity from different views, CV-EC can be formulated as follows:
\begin{equation}
\begin{split}
\mathcal{L}_{cv-ec}=\frac{1}{2M}\sum_{i=1}^{M}\frac{1}{K^1_iK^2_i}\sum_{p=1}^{K^1_i}\sum_{q=1}^{K^2_i}\parallel \bm{x}^1_{ip}-\bm{x}^2_{iq}\parallel^2_2.
\end{split}
\end{equation}
CV-EC aims to minimize the cross-view intra-class distances between the embedding feature pairs from different view-specific networks. We choose to implement the CV-EC metric between the last fully-connected layers. Thus, we can obtain view-specific information when extracting low-level features. The proposed framework jointly minimizes CV-EC and the view-specific softmax losses to extract discriminative features. The formula can be written as follows:
\begin{equation}
\begin{split}
\mathcal{L}_1=\sum_{v=1}^{2}\mathcal{L}_s^v+\lambda_1\mathcal{L}_{cv-ec},
\end{split}
\end{equation}
where $\mathcal{L}_s^v$ denotes the $v$th view-specific softmax loss and $\lambda_1$ represents the regularization factor between the softmax loss and CV-EC. The softmax loss \cite{wen2016discriminative} for the $v$th view can be formulated as follows:
\begin{equation}
\begin{split}
{\mathcal{L}}_s^v=-\sum_{n=1}^{N_v}log \frac{\mathrm{e}^{(\bm{W}^v_{y^v_n})^T\bm{x}^v_n+b^v_{y^v_n}}}{\sum_{m=1}^{M}\mathrm{e}^{(\bm{W}^v_m)^T\bm{x}^v_n+b_m^v}},
\end{split}
\end{equation}
\rev{where $\bm{x}^v_n$ denotes the $n$th deep feature from view $v$, belonging to the $y_n^v$th class; $\bm{W}_m^v$ represents the $m$th column of the weights $\bm{W^v}$ in the last fully-connected layer from view $v$; $b^v$ refers to the view-specific bias term; and $N_v$ indicates the size of the training samples in a mini-batch.}
\begin{figure}[t]
  \centering
  \includegraphics[width=0.48\textwidth]{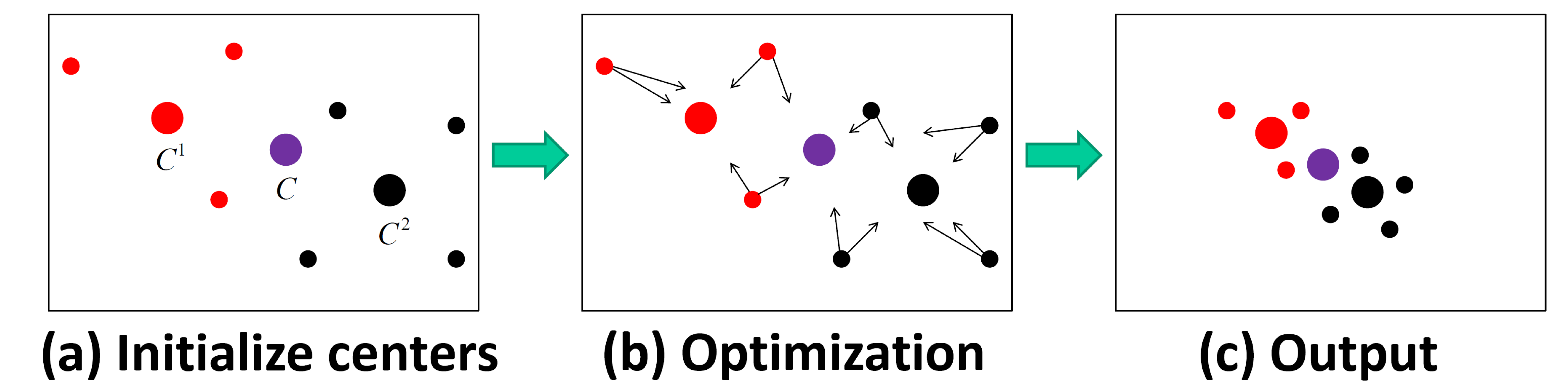}
  \caption{Cross-view center loss: (a) Initialize the centers for all the views and all the samples; (b) Optimization process that narrows the gaps between the samples and their corresponding centers; (c) Output result.}\label{3}
\end{figure}
\subsection{Cross-View Center Loss}
Center loss intends to learn discriminative features by penalizing the distances between the deep features and their corresponding centers. We can apply center loss to the re-id problem because each sequence captures multiple samples of the objective pedestrian. The original center loss \cite{wen2016discriminative} can be expressed as follows:
\begin{equation}
\begin{split}
\mathcal{L}_C=\frac{1}{2M}\sum_{i=1}^{M}\frac{1}{K_i}\sum_{j=1}^{K_i}\parallel \bm{x}_{ij}-\bm{C}_i\parallel^2_2,
\end{split}
\end{equation}
where $\bm{x}_{ij}$ denotes the $j$th deep feature of the $i$th person, $K_i$ refers to the number of the $i$th identity, and $\bm{C}_i$ indicates the $i$th class center.

View information is an intrinsic characteristic of re-id. When applying center loss to re-id, view information is useful for improving the performance. In this study, we extend center loss to a cross-view version (CV-CL). CV-CL is proposed to penalize the distances between the deep features and their corresponding view-specific centers. With CV-CL, we can narrow the gaps between different camera views. Figure 3 illustrates the idea of CV-CL.

Given training features $\{\bm{x}^1_{ip},\bm{x}^2_{iq}\}$, we can easily compute the centers and obtain $\{\bm{C}^1_{i},\bm{C}^2_{i},\bm{C}_i\}$, where $\bm{C}_{i}$ denotes the center of the $i$th class of all samples, and $\bm{C}^1_{i}$ and $\bm{C}^2_{i}$ represent the centers of the $i$th class of samples from the corresponding view. The formula for CV-CL is as follows:
\begin{equation}
\begin{split}
\mathcal{L}_{cv-cl}=\frac{1}{2M}\sum_{i=1}^{M}(\frac{1}{K^1_i}\sum_{p=1}^{K^1_i}(\parallel \bm{x}_{ip}^1-\bm{C}_i^1\parallel^2_2&+\parallel \bm{x}_{ip}^1-\bm{C}_i\parallel^2_2)\\
+\frac{1}{K^2_i}\sum_{q=1}^{K^2_i}(\parallel \bm{x}_{iq}^2-\bm{C}_i^2\parallel^2_2&+\parallel \bm{x}_{iq}^2-\bm{C}_i\parallel^2_2)).
\end{split}
\end{equation}
We combine CV-CL with the softmax loss. Therefore, the objective function can be written as follows:
\begin{equation}
\begin{split}
\mathcal{L}_2=\sum_{v=1}^{2}\mathcal{L}_s^v+\lambda_2\mathcal{L}_{cv-cl}.
\end{split}
\end{equation}
\subsection{Optimization of CV-EC and CV-CL}
\rev{ We adopt the standard stochastic gradient-based optimization algorithm to optimize the parameters of the view-specific networks by using CV-EC and CV-CL}. For CV-EC, the gradients of $\mathcal{L}_{cv-ec}$ with respect to $\bm{x}^1_{ip}$ and $\bm{x}^2_{iq}$ are computed as follows:
\begin{equation}
\begin{split}
\frac{\partial{\mathcal{L}_{cv-ec}}}{\partial{\bm{x}^1_{ip}}}=\bm{x}^1_{ip}-\bm{x}^2_{iq},\\
\frac{\partial{\mathcal{L}_{cv-ec}}}{\partial{\bm{x}^2_{iq}}}=\bm{x}^2_{iq}-\bm{x}^1_{ip}.
\end{split}
\end{equation}
\rev{By denoting the parameters of view-specific networks as $(\bm{\theta}^1,\bm{\theta}^2)$, we obtain the following:
\begin{equation}
\begin{split}
\frac{\partial{\mathcal{L}_1}}{\partial{\bm{\theta}^1}}=\frac{\partial{\mathcal{L}^1_s}}{\partial{\bm{\theta}^1}}+\
\frac{\partial{\mathcal{L}_{cv-ec}}}{\partial{\bm{x}^1_{ip}}}\cdot\frac{\partial{\bm{x}^1_{ip}}}{\partial{\bm{\theta}^1}},\\
\frac{\partial{\mathcal{L}_1}}{\partial{\bm{\theta}^2}}=\frac{\partial{\mathcal{L}^2_s}}{\partial{\bm{\theta}^2}}+\
\frac{\partial{\mathcal{L}_{cv-ec}}}{\partial{\bm{x}^2_{iq}}}\cdot\frac{\partial{\bm{x}^2_{iq}}}{\partial{\bm{\theta}^2}}.
\end{split}
\end{equation}
Finally, we update $(\bm{\theta}^1,\bm{\theta}^2)$ with the learning rate $\mu$ as follows:
\begin{equation}
\begin{split}
\bm{\theta}^1:=\bm{\theta}^1-\mu\cdot\frac{\partial{\mathcal{L}_1}}{\partial{\bm{\theta}^1}},\\
\bm{\theta}^2:=\bm{\theta}^2-\mu\cdot\frac{\partial{\mathcal{L}_1}}{\partial{\bm{\theta}^2}}.
\end{split}
\end{equation}}

For CV-CL, the gradients of $\mathcal{L}_{cv-cl}$ with respect to $\bm{x}^1_{ip}$ and $\bm{x}^2_{iq}$ can be obtained as follows:
\begin{equation}
\begin{split}
\frac{\partial{\mathcal{L}_{cv-cl}}}{\partial{\bm{x}^1_{ip}}}&=(\bm{x}^1_{ip}-\bm{C}^1_i)+(\bm{x}^1_{ip}-\bm{C}_i),\\
\frac{\partial{\mathcal{L}_{cv-cl}}}{\partial{\bm{x}^2_{iq}}}&=(\bm{x}^2_{iq}-\bm{C}^2_i)+(\bm{x}^2_{iq}-\bm{C}_i),\\
\end{split}
\end{equation}
The gradients of $\bm{C}^1_i$, $\bm{C}^2_i$, and $\bm{C}_i$ can be obtained as follows:
\begin{equation}
\begin{split}
\frac{\partial{\mathcal{L}_{cv-cl}}}{\partial{\bm{C}^1_i}}&=\bm{C}^1_i-\bm{x}^1_{ip},\\
\frac{\partial{\mathcal{L}_{cv-cl}}}{\partial{\bm{C}^2_i}}&=\bm{C}^2_i-\bm{x}^2_{iq},\\
\frac{\partial{\mathcal{L}_{cv-cl}}}{\partial{\bm{C}_i}}&=(\bm{C}_i-\bm{x}^1_{ip})+(\bm{C}_i-\bm{x}^2_{iq}).
\end{split}
\end{equation}
\rev{The parameters $(\bm{\theta}^1,\bm{\theta}^2)$ are updated as follows:
\begin{equation}
\begin{split}
\frac{\partial{\mathcal{L}_2}}{\partial{\bm{\theta}^1}}&=\frac{\partial{\mathcal{L}^1_s}}{\partial{\bm{\theta}^1}}+\
\frac{\partial{\mathcal{L}_{cv-cl}}}{\partial{\bm{x}^1_{ip}}}\cdot\frac{\partial{\bm{x}^1_{ip}}}{\partial{\bm{\theta}^1}},\\
\frac{\partial{\mathcal{L}_2}}{\partial{\bm{\theta}^2}}&=\frac{\partial{\mathcal{L}^2_s}}{\partial{\bm{\theta}^2}}+\
\frac{\partial{\mathcal{L}_{cv-cl}}}{\partial{\bm{x}^2_{iq}}}\cdot\frac{\partial{\bm{x}^2_{iq}}}{\partial{\bm{\theta}^2}},\\
\bm{\theta}^1:&=\bm{\theta}^1-\mu\cdot\frac{\partial{\mathcal{L}_2}}{\partial{\bm{\theta}^1}},\\
\bm{\theta}^2:&=\bm{\theta}^2-\mu\cdot\frac{\partial{\mathcal{L}_2}}{\partial{\bm{\theta}^2}}.
\end{split}
\end{equation}
Finally, we update $\{\bm{C}^1_{i},\bm{C}^2_{i},\bm{C}_i\}$ with the hyperparameter $\alpha$ as follows:
\begin{equation}
\begin{split}
\bm{C}^1_{i}:&=\bm{C}^1_{i}-\alpha\cdot\frac{\partial{\mathcal{L}_{cv-cl}}}{\partial{\bm{C}^1_{i}}},\\
\bm{C}^2_{i}:&=\bm{C}^2_{i}-\alpha\cdot\frac{\partial{\mathcal{L}_{cv-cl}}}{\partial{\bm{C}^2_{i}}},\\
\bm{C}_{i}:&=\bm{C}_{i}-\alpha\cdot\frac{\partial{\mathcal{L}_{cv-cl}}}{\partial{\bm{C}_{i}}}.
\end{split}
\end{equation}}
\subsection{Iterative Optimization}
Simultaneously learning CV-EC and CV-CL in a united framework is difficult. The optimization processes of CV-EC and CV-CL affect each other because of different convergence speeds. As the deep features from changing viewpoints are disparate, the joint optimization may become diverse with a large learning rate or result in a locally optimal output with a small learning rate. \rev{Further, balancing the view-specific networks and the proposed cross-view constraints in a united framework will be difficult because the search space of the regularization coefficients will increase quadratically.}

\rev{Training view-specific networks with either CV-EC or CV-CL separately is relatively easy. We can extract the discriminative features from the updated view-specific networks. With a substantially reduced cross-view intra-class distance, the view-specific deep networks optimized by CV-EC or CV-CL will be a better initialization for the other constraint.} On the basis of the above observations, we propose an iterative algorithm to update the parameters of the view-specific networks.

The iterative optimization algorithm is presented in Algorithm 1. In particular, the parameters of view-specific networks are first optimized by either CV-EC or CV-CL, and then the updated models are utilized as the initialization of the other cross-view constraint. Through iterative optimization, we can optimize the view-specific deep networks from coarse to fine.
\begin{algorithm} [t]
\caption{Iterative optimization algorithm}
\begin{algorithmic}[1]
\Require
Training samples $\{\bm{x}^1_{ip},\bm{x}^2_{iq}\}$, hyperparameter $\alpha$, $\lambda_1$, and $\lambda_2$, learning rate $\mu$;
\Ensure
Optimal view-specific parameters ${(\bm{\theta}^1,\bm{\theta}^2)}$;
\State Initialize the parameters of the view-specific networks with the pre-trained model on the re-id dataset;
\Repeat
\Repeat\\
\qquad Compute joint loss: $\mathcal{L}_1=\sum_{v=1}^{2}\mathcal{L}_s^v+\lambda_1\mathcal{L}_{cv-ec}$;\\
\qquad \rev{Compute $(\frac{\partial{\mathcal{L}_{cv-ec}}}{\partial{\bm{x}^1_{ip}}},\frac{\partial{\mathcal{L}_{cv-ec}}}{\partial{\bm{x}^2_{iq}}})$ for each i by Eq.(7);\\
\qquad Update ${(\bm{\theta}^1,\bm{\theta}^2)}$ using Eq.(8)-Eq.(9);}
\Until{$\mathcal{L}_1<\epsilon_1$};\\
\qquad Compute $(\bm{C}^1_i,\bm{C}^2_i,\bm{C}_i)$ for each i with ${(\bm{\theta}^1,\bm{\theta}^2)}$;
\Repeat\\
\qquad Compute joint loss: $\mathcal{L}_2=\sum_{v=1}^{2}\mathcal{L}_s^v+\lambda_2\mathcal{L}_{cv-cl}$;\\
\qquad \rev{Compute $(\frac{\partial{\mathcal{L}_{cv-cl}}}{\partial{\bm{x}^1_{ip}}},\frac{\partial{\mathcal{L}_{cv-cl}}}{\partial{\bm{x}^2_{iq}}})$ for each i by Eq.(10);\\
\qquad Update ${(\bm{\theta}^1,\bm{\theta}^2)}$ using Eq.(12);\\
\qquad Get $(\frac{\partial{\mathcal{L}_{cv-cl}}}{\partial{\bm{C}^1_i}},\frac{\partial{\mathcal{L}_{cv-cl}}}{\partial{\bm{C}^2_i}},\frac{\partial{\mathcal{L}_{cv-cl}}}{\partial{\bm{C}_i}})$ for each i by Eq.(11);\\
\qquad Update $(\bm{C}^1_i,\bm{C}^2_i,\bm{C}_i)$ for each i using Eq.(13);}
\Until{$\mathcal{L}_2<\epsilon_2$};\\
\qquad $\mathcal{L}=\mathcal{L}_1+\mathcal{L}_2$;
\Until{$\mathcal{L}<\epsilon$};
\end{algorithmic}
\end{algorithm}
\subsection{Multi-View CV-EC and CV-CL}
In real surveillance systems, a large number of cameras are placed in different positions. We need to develop an effective framework for multi-view re-id tasks. In the case of more than two camera views, a natural solution is to generate multiple view-specific networks. Then, we can learn a one-to-one cross-view constraint for each pair of the view-specific networks. However, the method mentioned above has two drawbacks. First, the simultaneous training of multiple deep networks requires a large amount of computational resources. The required resources increase proportionally to the number of cameras. Second, for a system with $N$ views, we need to optimize $C_N^2$ one-to-one objective loss functions during training. Optimizing a model with so many loss functions is difficult.

To solve this problem, we propose a simple but efficient iterative optimization algorithm. For each camera view, samples from all other views can be considered to be from $``$the other view$"$. Then, the multi-view re-id problem can be regarded as multiple one-to-others cross-view re-id problems. We can apply Algorithm 1 to solve these problems.

The multi-view optimization method is presented in Algorithm 2. For each view, we obtain two deep networks, one containing the view-generic information and the other containing the general information. We call the above networks the view-specific network and the public network, respectively. Then, the parameters of the public network are used for the initialization of the next iteration. Finally, we obtain all the view-specific networks.
\begin{algorithm} [t]
\caption{Multi-view optimization algorithm}
\begin{algorithmic}[1]
\Require
Multi-view deep features;
\Ensure
Optimal multi-view parameters ${(\bm{\theta}^1,...,\bm{\theta}^N,\bm{\theta})}$;
\State Initialize the view-specific networks and the public network with the pre-trained model, set $v=1$;
\Repeat
\Repeat\\
\qquad Update $(\bm{\theta}^v)$ and $(\bm{\theta})$ through CV-EC, loss $\mathcal{L}_1$;\\
\qquad Update $(\bm{\theta}^v)$ and $(\bm{\theta})$ through CV-CL, loss $\mathcal{L}_2$;\\
\qquad Compute total error: $\mathcal{L}=\mathcal{L}_1+\mathcal{L}_2$;
\Until{$\mathcal{L}<\epsilon$};\\
\qquad $v:=v+1$
\Until{$v>N$};
\end{algorithmic}
\end{algorithm}
\section{Experiment}
We have conducted extensive experiments on several public re-id datasets to validate the effectiveness of the proposed framework. In Section \uppercase\expandafter{\romannumeral4}-A, we introduce the datasets, the evaluation protocols, and the implementation details. In Section \uppercase\expandafter{\romannumeral4}-B, we compare the proposed approach with the state-of-the-art methods. In Section \uppercase\expandafter{\romannumeral4}-C, we evaluate the components of the proposed approach in detail.
\subsection{Experimental Settings}
\textbf{Datasets.} In this study, we evaluate the proposed method on five re-id benchmarks: VIPeR \cite{gray2007evaluating}, CUHK01 \cite{li2012human}, CUHK03 \cite{li2014deepreid}, SYSU-mReID \cite{guo2014multi}, and Market-1501 \cite{zheng2015scalable}. \emph{\textbf{VIPeR}} contains 1,464 images of 632 pedestrians captured in two camera views. This dataset is challenging for deep models because of the poor image quality and the lack of training samples. \emph{\textbf{CUHK01}} contains 3,884 images of 971 identities from two outdoor cameras. Each identity has two images per view. The image quality of the pedestrians in CUHK01 is better than that in VIPeR. \emph{\textbf{CUHK03}} is one of the largest re-id benchmarks in the existing literature. It contains more than 14,000 images of 1,467 people captured from six cameras. The images of each person are from two disjoint cameras. The pedestrians are cropped automatically or manually. \emph{\textbf{SYSU-mReid}} is customized for multi-shot re-id. It contains more than 24,000 images of 502 people from two cameras. SYSU-mReid is challenging for the existing methods. \emph{\textbf{Market-1501}} is a high-quality multi-view dataset for re-id. It contains 32,668 annotated bounding boxes of 1,501 identities from six cameras in an open system. The images are automatically detected by a deformable parts model detector.

\textbf{Evaluation protocols.} We use the standard protocol to ensure fair comparisons between the proposed method and the state-of-the-art methods. The test protocols are as follows. (\textbf{\uppercase\expandafter{\romannumeral1}}) For VIPeR, CUHK01, and SYSU-mReID benchmarks, we randomly split the datasets into two parts. Half the identities are used for training, while the rest identities are used for testing. The cumulative matching characteristic (CMC) is used to evaluate the performance of the compared methods. (\textbf{\uppercase\expandafter{\romannumeral2}}) For CUHK03, we follow the standard protocol used by \cite{chen2017person}: repeat 20 times to randomly split the samples into 100 people for testing and the remainder for training. We randomly select one image from the gallery for each identity and use all the images in the probe set to obtain the CMC curves. The evaluation process is repeated 100 times\footnote{\url{http://www.ee.cuhk.edu.hk/~xgwang/CUHK_identification.html}}, and the average value is computed as the final result. (\textbf{\uppercase\expandafter{\romannumeral3}}) For Market-1501, the standard protocol is defined by  \cite{zheng2015scalable}: train the models with the fixed training set (750 identities) and match 3,368 query images with the fixed testing set (751 identities). We conduct comparisons by using both single-query and multi-query (using multiple probe and query images) settings. Rank-1 accuracy and mean average precision (mAP) are computed to evaluate the performance of all the methods.
\begin{table}[t]
\centering
\caption{Details of Deep Networks}
\begin{tabular}{|c|c|c|c|}
\hline
\multicolumn{2}{|c|}{Alexnet}&\multicolumn{2}{c|}{JSTL\_DGD}\\
\hline
\hline
Layer name& Output size& Layer name& Output size\\
\hline
input& 227$\times$227$\times$3& input& 144$\times$56$\times$3\\
\hline
conv1& 55$\times$55$\times$96& conv1& 144$\times$56$\times$32\\
\hline
pool1& 27$\times$27$\times$96& conv2& 144$\times$56$\times$32\\
\hline
conv2& 27$\times$27$\times$256& conv3& 144$\times$56$\times$32\\
\hline
pool2& 13$\times$13$\times$256& pool3& 72$\times$28$\times$32\\
\hline
conv3& 13$\times$13$\times$384& inception (4a)& 72$\times$28$\times$256\\
\hline
conv4& 13$\times$13$\times$384& inception (4b)& 72$\times$28$\times$384\\
\hline
conv5& 13$\times$13$\times$256& inception (5a)& 36$\times$14$\times$512\\
\hline
pool5& 6$\times$6$\times$256& inception (5b)& 36$\times$14$\times$768\\
\hline
fc6& 4096& inception (6a)& 36$\times$14$\times$1024\\
\hline
fc7& 4096& inception (6b)& 36$\times$14$\times$1536\\
\hline
fc8& 512& fc7& 256\\
\hline
fc9& M& fc8& M\\
\hline
\end{tabular}
\end{table}

\rev{\textbf{Implementation details.} We adopt Alexnet \cite{krizhevsky2012imagenet} and JSTL\_DGD \cite{xiao2016learning} as the baseline models. The implementation details of Alexnet and JSTL\_DGD are presented in Table \uppercase\expandafter{\romannumeral1}. For Alexnet, we add an additional full connection layer before the softmax loss layer to form a deeper network. For JSTL\_DGD, we implement the baseline model with the source code available online\footnote{\url{https://github.com/Cysu/dgd_person_reid}}. Note that for the baseline models, a single network is trained and shared for disjoint views. The proposed methods are implemented using the Caffe CNN Library \cite{jia2014caffe}. The parameters are optimized in the stochastic gradient descent manner with a momentum of 0.9, weight decay of $10^{-4}$, learning rate $\mu$ of $10^{-4}$, and hyperparameter $\alpha$ of $10^{-3}$. We set the regularization coefficients $\lambda_1=\lambda_2=0.1$ for all the datasets. For Alexnet, the parameters are initialized with the model in \cite{krizhevsky2012imagenet}. For JSTL\_DGD, the parameters are initialized with the model in \cite{xiao2016learning}.}
\subsection{Comparisons with the State-of-the-Art Models}
In this section, we compare the proposed framework with the state-of-the-art approaches for both cross-view and multi-view ($>$2) re-id tasks. The evaluation was carried out on datasets from two cameras, such as VIPeR, CUHK01, and SYSU-mREID, and datasets from multiple cameras, such as CUHK03 and Market-1501.
\begin{table}[t]
\centering
\caption{Comparisons on VIPeR}
\begin{tabular}{|c|c|c|c|c|}
\hline
Rank (\%)& 1& 5& 10& 20\\
\hline
\hline
KISSME \cite{koestinger2012large}& 22& -& 68& -\\
\hline
rKPCCA \cite{mignon2012pcca}& 22.3& 55.5& 72.4& 86\\
\hline
LFDA \cite{pedagadi2013local}& 24.2& 52& 67.1& 82\\
\hline
CPDL \cite{li2015cross}& 34.0& 64.2& 77.5& 88.6\\
\hline
MLACM \cite{shi2015person}& 34.9& 59.3& 70.2& 81.8\\
\hline
Gated-SCNN \cite{varior2016gated}& 37.8& 66.9& 77.4& -\\
\hline
Deep Ranking \cite{chen2016deep}& 38.4& 69.2& 81.3& 90.4\\
\hline
XQDA \cite{liao2015person}& 40& 68.1& 80.5& 91.1\\
\hline
WARCA \cite{jose2016scalable}& 40.2& 68.2& 80.7& 91.1\\
\hline
S-LSTM \cite{varior2016siamese}& 42.4& 68.7& 79.4& -\\
\hline
HVIL-RMEL \cite{wang2016human}& 42.4& 72.6& 83& 90.4\\
\hline
CVDCA \cite{chen2016asymmetric}& 43.3& 72.7& 83.5& 92.2\\
\hline
Metric Ensemble \cite{paisitkriangkrai2015learning}& 45.9& 77.5& 88.9& 95.8\\
\hline
TCP \cite{cheng2016person}& 47.8& 74.7& 84.8& 89.2\\
\hline
FFN Net \cite{wu2016enhanced}& 51.1& 81& 91.4& 96.9\\
\hline
DNS \cite{zhang2016learning}& 51.2& 82.1& 90.5& 95.9\\
\hline
SSM \cite{bai2017scalable}& 50.73& -& 90.0& 95.6\\
\hline
SSM (Fusion) \cite{bai2017scalable}& 53.7& -& 91.5& 96\\
\hline
CRAFT-MFA \cite{chen2017person}& 50.3& 80.0& 89.6& 95.5\\
\hline
CRAFT-MFA (+LOMO) \cite{chen2017person}& \textbf{54.2}& \textbf{82.4}& \textbf{91.5}& \textbf{96.9}\\
\hline
JSTL\_DGD \cite{xiao2016learning}& 47.2& 73.4& 80.4& 86.4\\
\hline
\textbf{JSTL\_DGD+CV-EC}& 51.9& 76.6 & 85.4& 93.4\\
\hline
\end{tabular}
\end{table}

\textbf{(1) Comparisons on VIPeR.}
We compare the proposed approach with 18 methods on VIPeR, including metric learning-based models, view-specific models, and deep models. The details of the comparisons are presented in Table \uppercase\expandafter{\romannumeral2}. From Table \uppercase\expandafter{\romannumeral2}, we can see that the proposed view-specific deep network-based framework is competitive against other models. The experimental results validate that CV-EC is effective in improving the existing deep networks even without a sufficient number of training samples. The proposed approach manages to increase the rank-1 accuracy of JSTL\_DGD from 47.2\% to 51.9\%. Because only one sample per view is used for each identity, CV-CL and ICV-ECCL cannot be utilized for VIPeR. With an increasing number of training samples, we can implement view-specific networks with ICV-ECCL and achieve better recognition accuracy. The recognition accuracy of the proposed model is lower than that of CRAFT-MFA and SSM because of the lack of training samples. \rev{Moreover, both CRAFT-MFA and SSM benefit from the fusion of multiple features. Our work outperforms CRAFT-MFA by 1.6\% and SSM by 1.17\% in terms of the rank-1 accuracy calculated using the single features. We can further improve the performance of the view-specific networks by using the feature fusion strategy.}

\textbf{(2) Comparisons on CUHK01.} We manage to learn view-specific deep networks with ICV-ECCL on CUHK01. Table \uppercase\expandafter{\romannumeral3} shows the comparisons between the proposed framework and the other state-of-the-art methods. The proposed framework outperforms the compared methods. We notably improve the highest rank-1 accuracy from 78.8\% (by CRAFT-MFA) to 83.5\%. \rev{Further, our work outperforms CRAFT-MFA by 9\% in terms of the rank-1 accuracy calculated using the single features.} Compared with the original JSTL\_DGD model, a 7\% improvement in the rank-1 accuracy is observed. The proposed approach also improves the performance of Alexnet by 13.1\% in the rank-1 accuracy.

\begin{table}[t]
\centering
\caption{Comparisons on CUHK01}
\begin{tabular}{|c|c|c|c|c|}
\hline
Rank (\%)& 1& 5& 10& 20\\
\hline
\hline
LMNN \cite{weinberger2009distance}& 13.4& 31.3& 42.3& 54.1\\
\hline
ITML \cite{davis2007information}& 16& 35.2& 45.6& 59.8\\
\hline
rKPCCA \cite{mignon2012pcca}& 16.7& 41& 54.1& 67.7\\
\hline
Alexnet \cite{krizhevsky2012imagenet}& 18.7& 42.5& 55.6& 66.7\\
\hline
\textbf{Alexnet+ICV-ECCL}& 31.8& 58.4& 70.2& 80.6\\
\hline
XQDA \cite{liao2015person}& 63.2& 83.9& 90& 94.9\\
\hline
CVDCA \cite{chen2016asymmetric}& 47.8& 74.2& 83.4& 89.9\\
\hline
Deep Ranking \cite{chen2016deep}& 50.4& 75.9& 84& 91.3\\
\hline
Metric Ensemble \cite{paisitkriangkrai2015learning}& 53.4& 76.4& 84.4& 90.5\\
\hline
WARCA \cite{jose2016scalable}& 65.6& 85.3& 90.5& 95\\
\hline
TCP \cite{cheng2016person}& 53.7& 84.3& 91& 96.3\\
\hline
FFN Net \cite{wu2016enhanced}& 55.5& 78.4& 83.7& 92.6\\
\hline
DNS \cite{zhang2016learning}& 69.1& 86.9& 91.8& 95.4\\
\hline
CRAFT-MFA \cite{chen2017person}& 74.5& 91.2& 94.8& 97.1\\
\hline
CRAFT-MFA (+LOMO) \cite{chen2017person}& 78.8& 92.6& 95.3& 97.8\\
\hline
JSTL\_DGD \cite{xiao2016learning}& 76.5& 92.4& 95.3& 97.3\\
\hline
\textbf{JSTL\_DGD+ICV-ECCL}& \textbf{83.5}& \textbf{95.2} & \textbf{97.3}& \textbf{98.8}\\
\hline
\end{tabular}
\end{table}
\begin{table}[t]
\centering
\caption{Comparisons on CUHK03}
\begin{tabular}{|c|c|c|c|c|}
\hline
Rank (\%)& 1& 5& 10& 20\\
\hline
\hline
LMNN \cite{weinberger2009distance}& 5.1& 17.7& 28.3& -\\
\hline
ITML \cite{davis2007information}& 6.3& 18.7& 29& -\\
\hline
KISSME \cite{koestinger2012large}& 11.7& 33.3& 48& -\\
\hline
BoW \cite{zheng2015scalable}& 23& 42.4& 52.4& 64.2\\
\hline
XQDA \cite{liao2015person}& 46.3& 78.9& 88.6& 94.3\\
\hline
HVIL \cite{wang2016human}& 56.1& 64.7& 75.7& 87.4\\
\hline
CVDCA \cite{chen2016asymmetric}& 47.8& 74.2& 83.4& 89.9\\
\hline
Metric Ensemble \cite{paisitkriangkrai2015learning}& 62.1& 89.1& 94.3& 97.8\\
\hline
S-LSTM \cite{varior2016siamese}& 57.3& 80.1& 88.3& -\\
\hline
WARCA \cite{jose2016scalable}& 75.4& 94.5& 97.5& 99.1\\
\hline
Alexnet \cite{krizhevsky2012imagenet}& 52.4& 84.8& 92.9& 97.4\\
\hline
\textbf{Alexnet+ICV-ECCL}& 69.9& 91.8& 97.3& 99.1\\
\hline
TCP \cite{cheng2016person}& 53.7& 84.3& 91& 96.3\\
\hline
DNS \cite{zhang2016learning}& 54.7& 84.8& 94.8& 95.2\\
\hline
Deep Histogram Loss \cite{ustinova2016learning}& 65.8& 92.9& 97.6& 99.4\\
\hline
SSM (Fusion) \cite{bai2017scalable}& 72.7& 92.4& 96& -\\
\hline
CRAFT-MFA \cite{chen2017person}& 84.3& 97.1& 98.3& 99.1\\
\hline
CRAFT-MFA (+LOMO) \cite{chen2017person}& 87.5& 97.4& 98.7& 99.5\\
\hline
JSTL\_DGD \cite{xiao2016learning}& 83.4& 97.1& 98.7& 99.5\\
\hline
\textbf{JSTL\_DGD+ICV-ECCL}& \textbf{88.6}& \textbf{98.2} & \textbf{99.2}& \textbf{99.7}\\
\hline
\end{tabular}
\end{table}
\textbf{(3) Comparisons on CUHK03.} Deep networks can achieve excellent performance with a sufficient number of training samples from CUHK03. Table \uppercase\expandafter{\romannumeral4} shows the comparisons of the proposed method and the other approaches. ICV-ECCL achieves the highest rank-1 accuracy and surpasses the best alternative (CRAFT-MFA) by 1.1\%. \rev{In particular, ICV-ECCL outperforms CRAFT-MFA by 4.3\% in terms of the rank-1 accuracy calculated using the single features.} A significant improvement is also observed in the rank-1 accuracy of Alexnet from 52.4\% to 69.9\%. The experimental results prove that exploiting view information during feature extraction is effective for multi-view re-id tasks.

\textbf{(4) Comparisons on SYSU-mREID.} Table \uppercase\expandafter{\romannumeral5} shows the experimental results for SYSU-mREID. The proposed framework remarkably outperforms the state-of-the-art methods with a margin of 23.3\% (64.1\%-40.8\%). With ICV-ECCL, the rank-1 accuracy of Alexnet/JSTL\_DGD is notably improved from 35\%/58.5\% to 42\%/65.1\%. The experimental results indicate that the proposed framework is also effective for multi-shot re-id tasks.
\begin{table}[t]
\centering
\caption{Comparisons on SYSU-mREID }
\begin{tabular}{|c|c|c|c|c|}
\hline
Rank (\%)& 1& 5& 10& 20\\
\hline
\hline
KISSME \cite{koestinger2012large}& 16.9& 39.8& 54.7& 69\\
\hline
rKPCCA \cite{mignon2012pcca}& 22.3& 53& 68.1& 83.7\\
\hline
LFDA \cite{pedagadi2013local}& 26.2& 55.6& 68.8& 80.3\\
\hline
MLACM \cite{shi2015person}& 30.8& 55.5& 67.4& 77\\
\hline
CVDCA \cite{chen2016asymmetric}& 40.8& 71.4& 82.2& 90.6\\
\hline
Alexnet \cite{krizhevsky2012imagenet}& 35& 62& 73.7& 84.3\\
\hline
\textbf{Alexnet+ICV-ECCL}& 42& 70& 80.4& 89\\
\hline
JSTL\_DGD \cite{xiao2016learning}& 58.5& 82.4& 87.7& 93.9\\
\hline
\textbf{JSTL\_DGD+ICV-ECCL}& \textbf{65.1}& \textbf{86.4} & \textbf{91.7}& \textbf{95.4}\\
\hline
\end{tabular}
\end{table}
\begin{table}[t]
\centering
\caption{Comparisons on Market-1501}
\begin{tabular}{|c|c|c|c|c|}
\hline
Query &\multicolumn{2}{|c|}{Single Query}&\multicolumn{2}{|c|}{Multiple Query}\\
\hline
Evaluation models & Rank-1 & mAP & Rank-1 & mAP \\
\hline
\hline
KISSME \cite{koestinger2012large}& 40.5& 19& -& -\\
\hline
BoW \cite{zheng2015scalable}& 34.4& 14.1& -& -\\
\hline
XQDA \cite{liao2015person}& 43.8& 22.2& 54.1& 28.4\\
\hline
WARCA \cite{jose2016scalable}& 45.2& -& -& -\\
\hline
S-LSTM \cite{varior2016siamese}& -& -& 61.6& 35.3\\
\hline
Deep Histogram Loss \cite{ustinova2016learning}& 59.5& -& -& -\\
\hline
Gated-SCNN \cite{varior2016gated}& 65.9& 39.6& 76& 48.5\\
\hline
DNS \cite{zhang2016learning}& 61& 35.7& 71.6& 46\\
\hline
IDE (R) \cite{zhong2017re}& 77.1& 63.6& -& -\\
\hline
SSM (Fusion) \cite{bai2017scalable}& 82.2& 68.8& 88.2& 76.2\\
\hline
CRAFT-MFA \cite{chen2017person}& 68.7& 42.3& 77.0& 50.3\\
\hline
CRAFT-MFA (+LOMO) \cite{chen2017person}& 71.8& 45.5& 79.7& 54.3\\
\hline
JSTL\_DGD \cite{xiao2016learning}& 81.6& 56.6& 83.1& 68.1\\
\hline
\textbf{JSTL\_DGD+ICV-ECCL}& \textbf{88.4}& \textbf{69.5} & \textbf{90.6}& \textbf{77.3}\\
\hline
\end{tabular}
\end{table}
\begin{table*}[t]
\newcommand{\tabincell}[2]{\begin{tabular}{@{}#1@{}}#2\end{tabular}}
\centering
\caption{Evaluation of the Overall Framework}
\scalebox{0.9}{
\begin{tabular}{|c|c|c|c|c|c|c|c|c|c|c|c|c|c|c|c|c|c|c|c|c|}
\hline
Dataset& \multicolumn{4}{|c|}{VIPeR}& \multicolumn{4}{|c|}{CUHK01}& \multicolumn{4}{|c|}{CUHK03}& \multicolumn{4}{|c|}{SYSU-mREID}& \multicolumn{4}{|c|}{Market-1501}\\
\hline
Rank (\%)& 1& 5& 10& 20& 1& 5& 10& 20& 1& 5& 10& 20& 1& 5& 10& 20& 1& 5& 10& 20\\
\hline
\hline
Alexnet \cite{krizhevsky2012imagenet}& -& -& -& -& 18.7& 42.5& 55.6& 66.7& 52.4& 84.8& 92.9& 97.4& 34.9& 62.0& 73.7& 84.3& -& -& -& -\\
\hline
\textbf{\tabincell{c}{Alexnet+\\ICV-ECCL}}& -& -& -& -& 31.8& 58.4& 70.2& 80.6& 69.9& 91.8& 97.3& 99.1& 42& 70& 80.4& 89& -& -& -& -\\
\hline
JSTL\_DGD \cite{xiao2016learning}& 47.2& 73.4& 80.4& 86.4& 76.5& 92.4& 95.3& 97.3& 83.4& 97.1& 98.7& 99.5& 58.5& 82.4& 89.2& 93.9& 81.6& 92.3& 95.3& 97\\
\hline
\textbf{\tabincell{c}{JSTL\_DGD+\\ICV-ECCL}}& \textbf{51.9}& \textbf{76.6}& \textbf{85.4}& \textbf{93.4}& \textbf{83.5}& \textbf{95.2}& \textbf{97.6}& \textbf{98.8}& \textbf{88.6}& \textbf{98.2}& \textbf{99.3}& \textbf{99.7}& \textbf{65.1}& \textbf{86.4}& \textbf{91.7}& \textbf{95.4}& \textbf{88.4}& \textbf{94.8}& \textbf{96.5}& \textbf{98}\\
\hline
\end{tabular}}
\end{table*}

\textbf{(5) Comparisons on Market-1501.} The experimental results for Market-1501 are presented in Table \uppercase\expandafter{\romannumeral6}. The proposed framework remains competitive for a realistic protocol. ICV-ECCL achieves the highest rank-1 accuracy and mAP for both single-query and multi-query re-id tasks. The proposed framework outperforms the best model (SSM) in the existing literature by 6.2\%/0.7\% for rank-1/mAP in the single-query task and 2.4\%/1.1\% for rank-1/mAP in the multi-query task. For the baseline model (JSTL\_DGD), ICV-ECCL exhibits an improvement of 6.8\%/13.3\% in rank-1/mAP in the single-query task and 7.5\%/9.2\% in rank-1/mAP in the multi-query task. The experimental results validate the effectiveness of the proposed approach in the case of a realistic testing protocol.

\textbf{(6) Conclusion of comparisons.} From the experimental results, we can conclude the following. First, learning view-specific deep networks with cross-view constraints can significantly improve the performance of the existing deep networks. ICV-ECCL is proved to be effective in enhancing the discriminative abilities of both Alexnet and JSTL\_DGD. Second, the proposed framework can be adapted to various re-id benchmarks. Comparisons on VIPeR and CUHK01 validate that view-specific deep networks are competitive even without a sufficient number of training samples. With enough training data, a significant improvement can be observed on CUHK03. The proposed approach is also adaptive to the multi-shot re-id task on SYSU-mREID and the multi-view realistic re-id task on Market-1501. Third, the proposed framework outperforms the state-of-the-art methods with respect to several re-id benchmarks, including the CUHK01, CUHK03, SYSU-mREID, and Market-1501 datasets.

\subsection{In-Depth Analysis of the Proposed Framework}
In this section, we conduct a quantitative self-evaluation to verify the effectiveness of the proposed framework in detail. We not only evaluate the effect of our overall framework but also assess each component to measure its contribution.

\begin{table}[t]
\centering
\caption{Performance with respect to different constraints}
\begin{tabular}{|c|c|c|}
\hline
Compared methods& Accuracy & Cross-view intra-class distance\\
\hline
\hline
Alexnet& 52.4& $1.27\times10^5$\\
\hline
Alexnet+CV-EC& 66.56& $3.74\times10^3$\\
\hline
Alexnet+CV-CL& 64.72& $4.81\times10^3$\\
\hline
Alexnet+Triplet& 60.8& -\\
\hline
Alexnet+ICV-ECCL& 69.94& $1.93\times10^3$\\
\hline
\end{tabular}
\end{table}

\textbf{Effectiveness of CV-EC and CV-CL.} We train Alexnet with CV-EC and CV-CL on CUHK03 to evaluate their contributions. CV-EC and CV-CL are compared with the baseline model and the triplet loss \cite{schroff2015facenet}. The experimental results are presented in Table \uppercase\expandafter{\romannumeral8}. We can see that CV-EC and CV-CL surpasses the baseline model by 14.16\% and 12.32\%, respectively. The cross-view intra-class distance is reduced to $3.74\times10^3$ by CV-EC and $4.81\times10^3$ by CV-CL from $1.27\times10^5$ of the baseline model. Moreover, CV-EC/CV-CL outperforms the triplet loss by 5.75\%/3.92\%. The above observation indicates that the proposed framework is more effective than the state-of-the-art deep metric learning methods for the re-id task. Note that CV-EC minimizes the gaps between different view-specific features, while CV-CL learns the concentrated clusters around the centers. Therefore, CV-EC outperforms CV-CL in terms of both the recognition accuracy and the cross-view intra-class distance.

\textbf{Effectiveness of the iterative optimization algorithm.} Table \uppercase\expandafter{\romannumeral9} shows the effect of the iterative learning algorithm. We can see that the recognition accuracy keeps increasing and the cross-view intra-class distance keeps decreasing during training. With iterative optimization, we obtain a better initialization for the view-specific networks after each step. The optimization process is thus from coarse to fine. Finally, we can obtain the desirable view-specific deep networks after several iterations.

\rev{\textbf{Effectiveness of the overall framework.} By comparisons with the baseline models, we quantitatively evaluate the effectiveness of the entire framework for the existing deep networks on various re-id benchmarks. The proposed framework is implemented over Alexnet and JSTL\_DGD and tested on the VIPeR, CUHK01, CUHK03, SYSU-mREID, and Market-1501 datasets. The experimental results are presented in Table \uppercase\expandafter{\romannumeral7}. A remarkable improvement compared with the baseline models is observed in the cases of all the datasets. For Alexnet, a performance gain of 17.5\%/13.1\%/7.02\% in the rank-1 accuracy is observed for the CUHK03/CUHK01/SYSU-mREID dataset. For JSTL\_DGD, a notable improvement of 4.75\%/7.05\%/5.23\%/6.64\%/6.83\% in the rank-1 accuracy is obtained for the VIPeR/CUHK01/CUHK03/SYSU-mREID/Market-1501 dataset.}
\begin{table}[t]
\centering
\caption{Performance with respect to iterative optimization}
\begin{tabular}{|c|c|c|}
\hline
Iteration step& Accuracy & Cross-view intra-class distance\\
\hline
\hline
Initial baseline& 52.4& $1.27\times10^5$\\
\hline
1st iter CV-EC& 66.56& $3.74\times10^3$\\
\hline
1st iter CV-CL& 68.12& $2.82\times10^3$\\
\hline
2nd iter CV-EC& 69.33& $1.96\times10^3$\\
\hline
2nd iter CV-CL& 69.94& $1.93\times10^3$\\
\hline
\end{tabular}
\end{table}
\begin{figure}[t]
  \centering
  \includegraphics[width=0.5\textwidth]{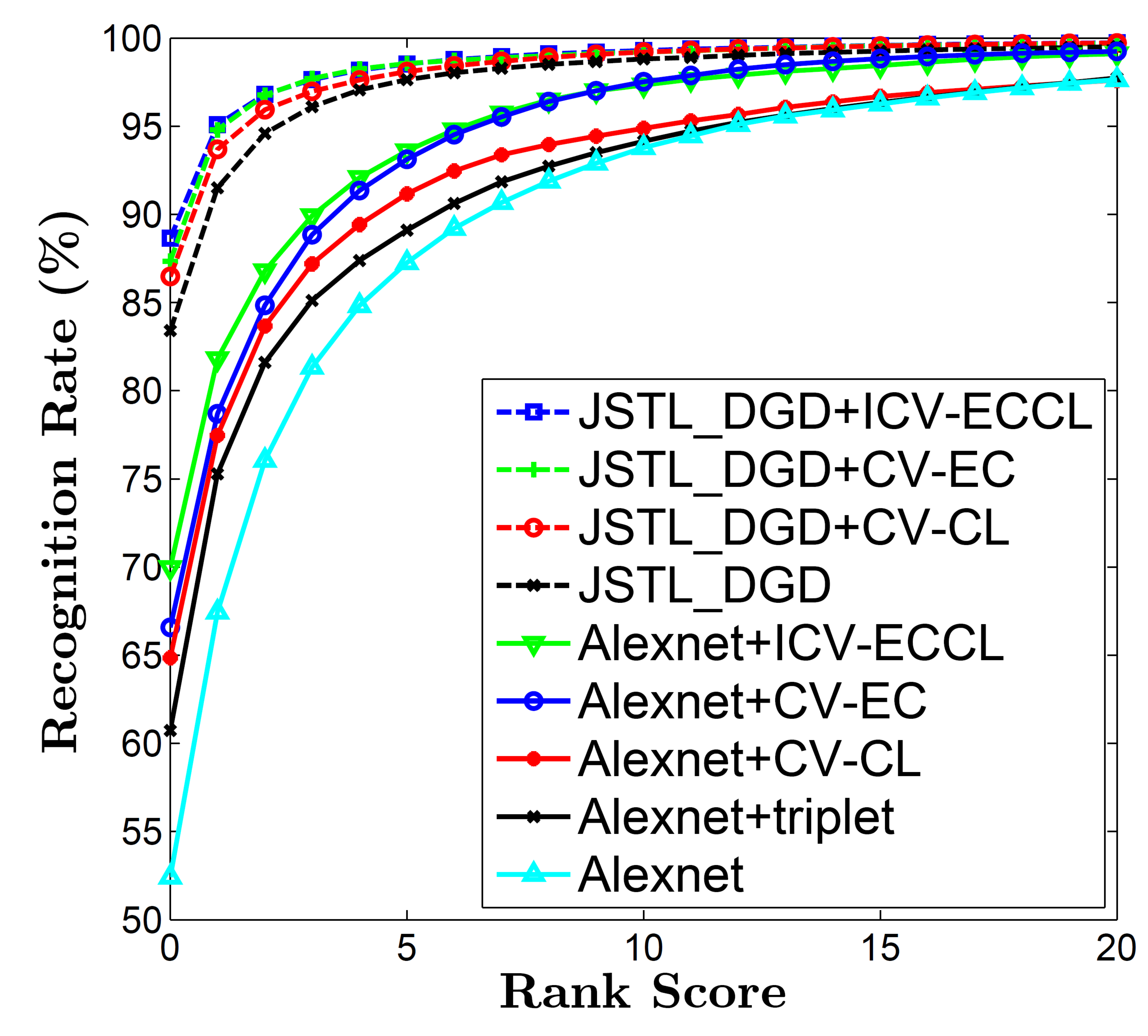}
  \caption{Evaluation of the proposed framework on CUHK03.}\label{7}
\end{figure}
Figure 5 shows the CMC curves of the overall framework and each component of the proposed approach on CUHK03. ICV-ECCL is clearly effective in improving the performance of Alexnet and JSTL\_DGD. Compared with the baseline model, a significant improvement from each component of the proposed approach is also observed.
\begin{figure} [t]
\begin{minipage}[t]{0.481\linewidth}
\centering
\includegraphics[width=\textwidth]{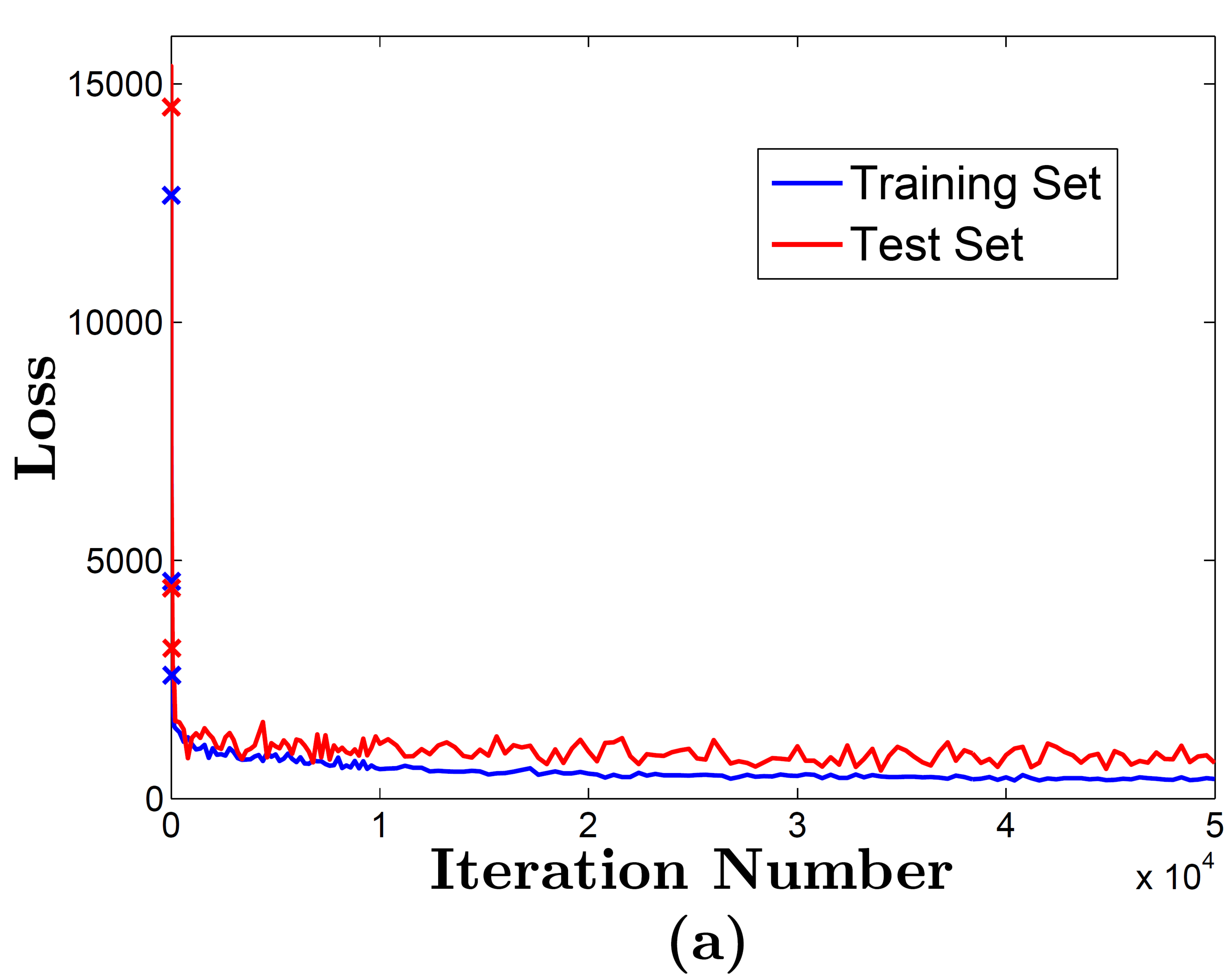}
\label{fig:side:a}
\end{minipage}%
\begin{minipage}[t]{0.48\linewidth}
\centering
\includegraphics[width=\textwidth]{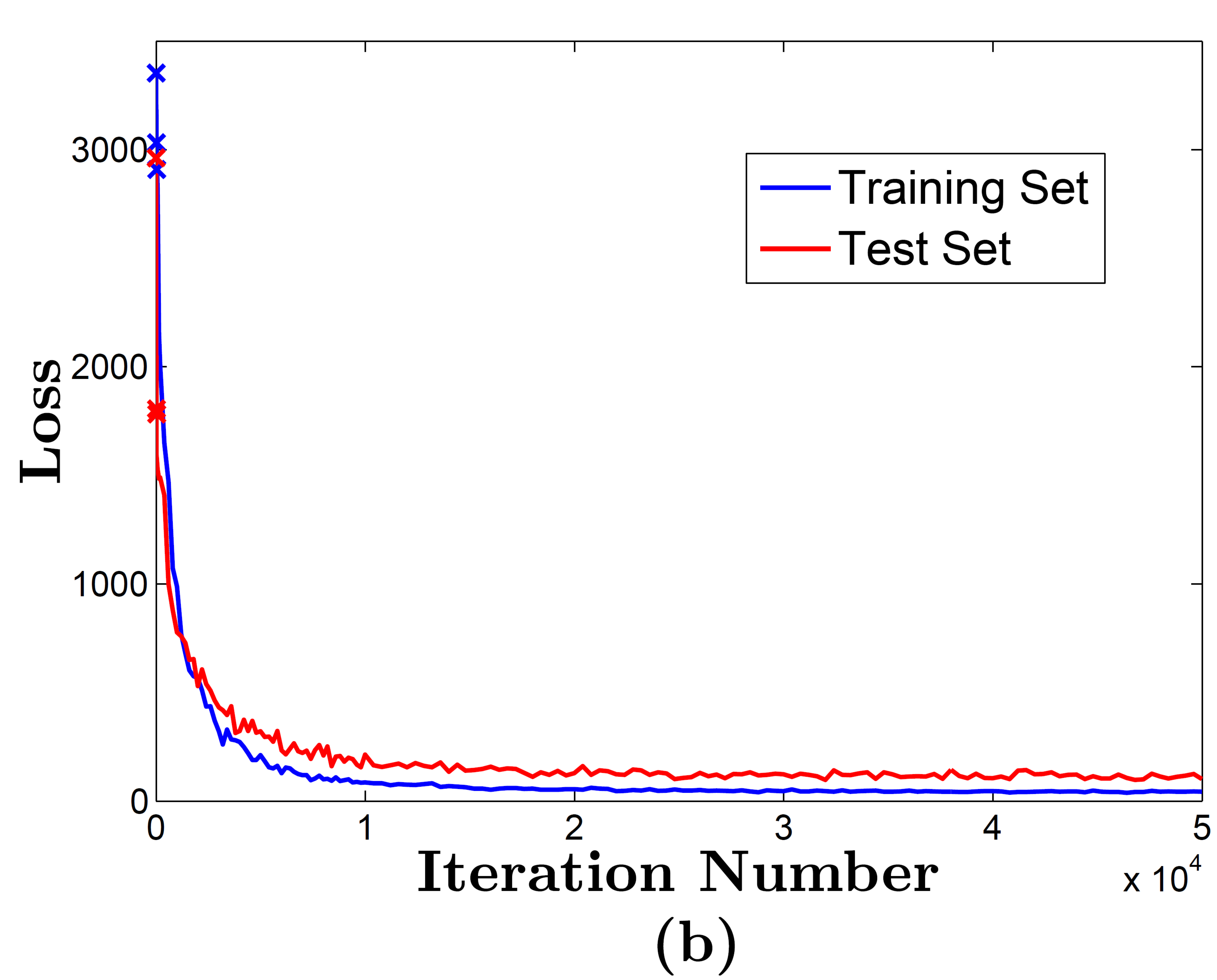}
\label{fig:side:b}
\end{minipage}
\caption{\rev{The convergence of CV-EC and CV-CL.} \rev{Figure 6(a) shows the training error and the test error of CV-EC, while Figure 6(b) shows the training error and the test error of CV-CL. The training loss and the test loss drop dramatically in the early iterations and keep decreasing for both CV-EC and CV-CL.}}
\end{figure}

\rev{\textbf{Evaluation of the convergence of CV-EC and CV-CL.} Figure 6 validates the convergence of CV-EC and CV-CL. We integrate CV-EC and CV-CL with Alexnet and train the view-specific networks on CUHK03. In Figure 6(a), we observe a fast convergence of CV-EC. The initial training loss from the baseline model is huge and drops dramatically after a few iterations. During training, the losses of both the training set and the test set keep decreasing. These decreases imply that the training process is convergent and the view-specific networks have excellent generalizability. Figure 6(b) illustrates the training process of CV-CL. The convergence of CV-CL is similar to that of CV-EC. Further, the training loss keeps decreasing along with the test loss. In the early stage of the optimization, the convergence speed of CV-CL is slower than that of CV-EC because more parameters are involved in CV-CL. Note that in the case of a small test loss, the generalizability of view-specific features generated by CV-CL is excellent.}

\textbf{Discussion on the regularization coefficients $\lambda_1$ and $\lambda_2$.} Figure 7(a) shows the relationship between the rank-1 accuracy of CV-EC and the regularization coefficient $\lambda_1$. $\lambda_1$ is used to balance CV-EC and softmax. We can see that CV-EC achieves the best recognition accuracy when $\lambda_1=0.1$. Therefore, we set $\lambda_1=0.1$ for all the datasets. Similarly, Figure 7(b) shows that CV-CL achieves the best recognition accuracy when $\lambda_2=0.1$. As with $\lambda_1$, we set $\lambda_2=0.1$ for all the datasets.

\rev{Figure 7 shows that the performance of both CV-EC and CV-CL remains relatively stable across a wide range of regularization coefficients (from $10^{-3}$ to $10^1$). We can conclude that implementing view-specific networks with cross-view constraints can improve the performance of deep features for most cases. Note that when $\lambda_1=\lambda_2=0$, the cross-view constraints will be removed. Without the cross-view constraints, the performance of the view-specific networks will decrease significantly. In contrast, when the regularization coefficients are infinitely large, the view-specific softmax losses will be suppressed. The deep features extracted by the view-specific networks may manage to minimize the view discrepancies but fail to distinguish the identities. Finally, we also consider the relationship between the regularization coefficients of CV-EC and CV-CL. The regularization coefficients are independent because the proposed framework adopts an iterative optimization algorithm. This framework can achieve the best performance with the best alternative regularization coefficients of CV-EC and CV-CL.}
\begin{figure} [t]
\begin{minipage}[t]{0.5\linewidth}
\centering
\includegraphics[width=\textwidth]{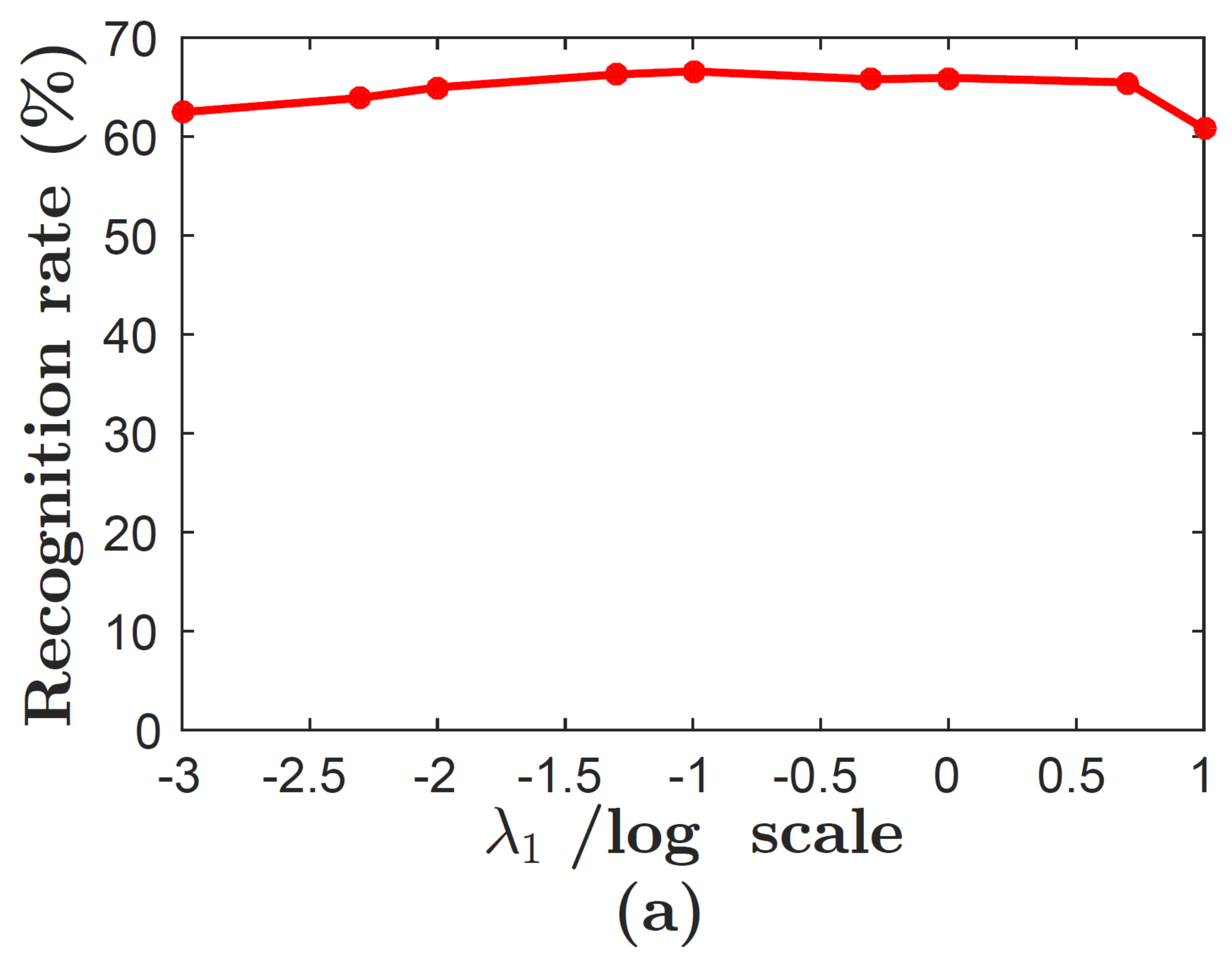}
\label{fig:side:a1}
\end{minipage}%
\begin{minipage}[t]{0.5\linewidth}
\centering
\includegraphics[width=\textwidth]{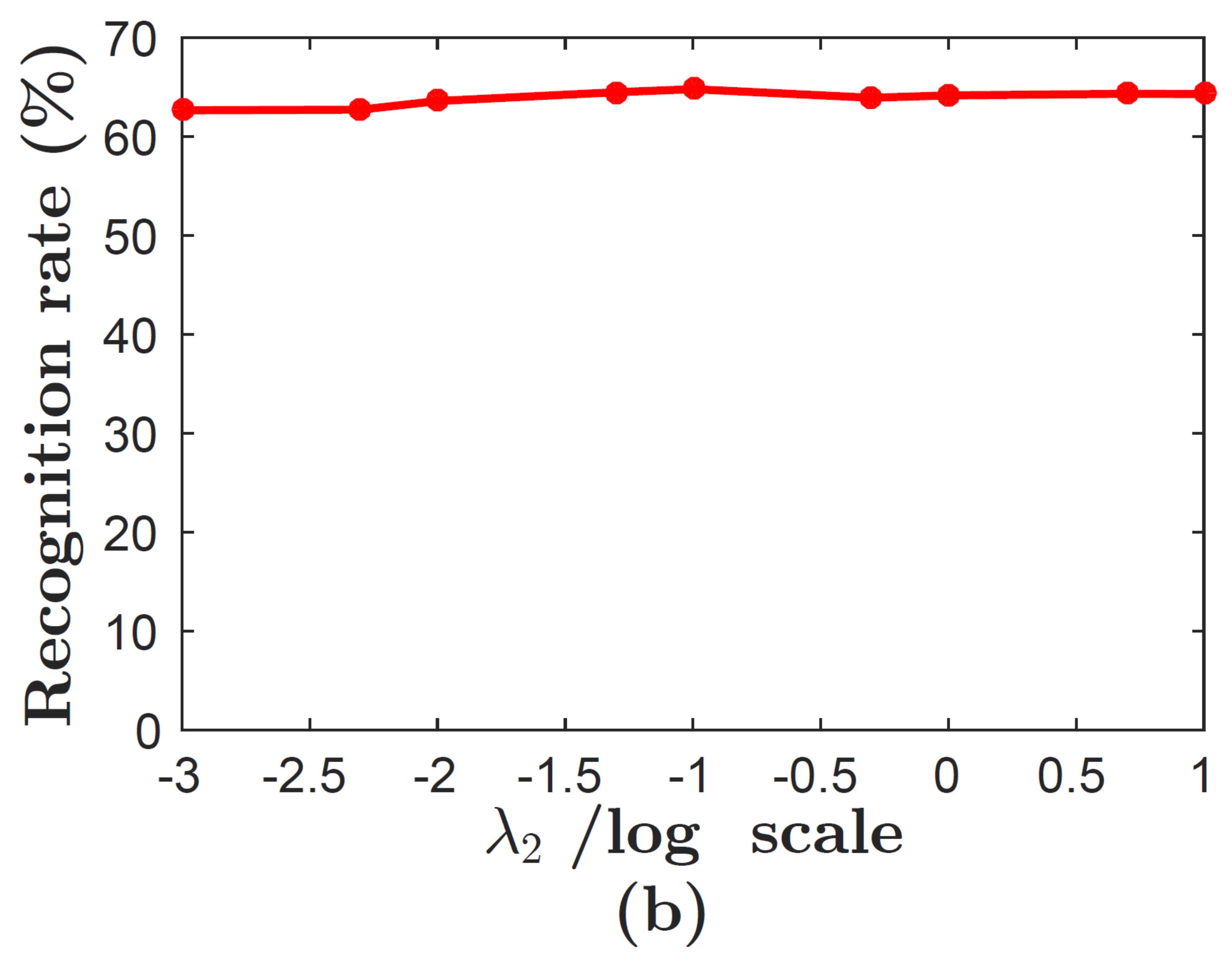}
\label{fig:side:b1}
\end{minipage}
\caption{Influence of $\lambda_1$ and $\lambda_2$ on the CUHK03 Dataset. Best performance is obtained when $\lambda_1=0.1$ and $\lambda_2=0.1$.}
\end{figure}

\section{Conclusion}
Our research focuses on learning view-specific deep networks for re-id, which has been overlooked in the existing literature. We propose a novel framework called the iterative cross-view Euclidean constraint and center loss (ICV-ECCL). The proposed framework exploits view-specific information during low-level feature extraction to minimize the cross-view intra-class distance. Moreover, ICV-ECCL adopts an iterative optimization algorithm to iteratively optimize the parameters of the view-specific networks by using CV-EC and CV-CL. We further extend ICV-ECCL to a multi-view version to cope with benchmarks captured from multiple camera views.

The proposed framework is implemented over Alexnet and JSTL\_DGD. We evaluate the effectiveness of ICV-ECCL on the VIPeR, CUHK01, CUHK03, SYSU-mREID, and Market-1501 benchmarks. The experiments validate that the proposed approach significantly improves the performance of both Alexnet and JSTL\_DGD in all the datasets. Moreover, the proposed framework outperforms the state-of-the-art methods on the CUHK01, CUUK03, SYSU-mREID, and Market-1501 datasets. We can conclude that learning view-specific deep networks with ICV-ECCL is effective for the re-id task.

\section*{Acknowledgment}
This project was supported by the NSFC (U1611461, 61573387, 61672544).

\ifCLASSOPTIONcaptionsoff
  \newpage
\fi



\end{document}